\documentclass[a4paper,12pt]{article}
\usepackage{graphicx}

\def\pslibdirtempcoh{figures}

\begin{document} 
\date{August 2000}
\title{\sf PROBABILISTIC  MOTION ESTIMATION BASED ON TEMPORAL COHERENCE}
\author{Pierre-Yves Burgi$^{1}$, Alan L. Yuille$^{2,\ast}$, and Norberto M. Grzywacz$^{2}$\\
    \\$^{1}$Centre Suisse d'Electronique et Microtechnique,\\ Jaquet-Droz 1, 2007 Neuch\^{a}tel, Switzerland,\\
     \\$^{2}$The Smith-Kettlewell Eye Research Institute,\\ 2232 
     Webster Street, San Francisco, CA 94115,\\
     \\$^{\ast}$To whom correspondence should be addressed.}

\maketitle

\newpage
\setcounter{page}{1}

\section*{Abstract}

We develop a theory for the temporal integration of visual motion
motivated by psychophysical experiments. The theory proposes that
input data are temporally grouped and used to predict and estimate the
motion flows in the image sequence. This temporal grouping can be
considered a generalization of the data association techniques used by
engineers to study motion sequences. Our temporal-grouping theory is
expressed in terms of the Bayesian generalization of standard Kalman
filtering. To implement the theory we derive a parallel network which
shares some properties of cortical networks. Computer simulations of
this network demonstrate that our theory qualitatively accounts for
psychophysical experiments on motion occlusion and motion outliers.

\newpage

\section{Introduction}
\label{Introduction}

Local motion signals are often ambiguous and many important motion
phenomena can be explained by hypothesising that the human visual
system uses temporal coherence to resolve ambiguous
inputs\footnote{Temporal coherence is the assumption that motion in
natural images is mostly temporally smooth, that is, it rarely changes
direction and/or speed abruptly.}.  For example, the perceptual
tendency to disambiguate the trajectory of an ambiguous motion path by
using the time average of its past motion was first reported by Anstis
and Ramachandran (Ramachandran and Anstis 1983; Anstis and
Ramachandran 1987). They called this phenomenon {\it motion
inertia}. Other motion phenomena involving temporal coherence include
the improvement of velocity estimation over time (McKee {\it et al.}
1986, Ascher 1998), blur removal (Burr {\it et al.} 1986),
motion-outlier detection (Watamaniuk {\it et al.}  1994), and motion
occlusion (Watamaniuk and McKee 1995). These motion phenomena pose a
serious challenge to motion-perception models. For example, in
motion-outlier detection the target dot is indistinguishable from the
background noise if one only observes a few frames of the motion.
Therefore, the only way to detect the target dot is by means of its
extended motion over time. Consequently, explanations based solely on
single, large local-motion detectors as in the Motion-Energy and
Elaborated-Reichardt models (Hassenstein and Reichardt 1956; van
Santen and Sperling 1984; Adelson and Bergen 1985; Watson and Ahumada
1985; for a review of motion models, see Nakayama 1985) seem
inadequate to explain these phenomena (Verghese {\it et al.}
1999). Also, it has been argued (Yuille and Grzywacz 1998) that all
these experiments could be interpreted in terms of temporal grouping
of motion signals involving prediction, observation and estimation.
The detection of targets, and their temporal grouping, could be
achieved by verifying that the observations were consistent with the
motion predictions. Conversely, failure in these predictions would
indicate that the observations were due to noise, or distractors,
which could be ignored. The ability of a system to predict the
target's motions thus represents a powerful mechanism to extract
targets from background distractors and might be a powerful cue for
the interpretation of visual scenes.

There has been little theoretical work on temporal coherence. Grzywacz
{\it et al.} (1989) developed a theoretical model which could account
for motion-inertia phenomena by requiring that the direction of motion
varies slowly over time while speed could vary considerably faster.
More recently Grzywacz {\it et al.}  (1995) proposed a biologically
plausible model that extends the work on motion inertia and allow for
the detection of motion outliers. It was observed (Grzywacz {\it et
al} 1989) that for general three-dimensional motion, the direction,
but not the speed, of the image motion is more likely to be constant
and hence might be more reliable for motion prediction. This is
consistent with the psychophysical experiments on motion inertia
(Anstis and Ramachandran, 1987) and temporal smoothing of motion
signals (Gottsdanker 1956; Snowden and Braddick 1989; Werkhoven {\it
et al.}  1992) which demonstrated that the direction of motion is the
primary cue in temporal motion coherence. Other reasons which might
make motion direction more reliable than speed are the unreliability
of local-speed measurement due to the poor temporal resolution of
local motion units (Kulikowski and Tolhurst, 1973, Holub and
Morton-Gibson 1981) and to involuntary eye movements.

The aim of this paper is to present a new theory for visual-motion
estimation and prediction that exploits temporal information. This
theory builds on our previous work (Grzywacz {\it et al.} 1989;
Grzywacz {\it et al.} 1995; Yuille {\it et al.} 1998) and on recent
work on tracking of object boundaries (Isard and Blake 1996). In
addition, this paper gives a concrete example of the abstract
arguments on temporal grouping proposed by Yuille and Grzywacz
(1998). To this end, we use a Bayesian formulation of estimation over
time (Ho and Lee 1964), which allows us simultaneously to make
predictions over time, update those predictions using new data, and
reject data that are inconsistent with the predictions\footnote{We
will be updating the {\it marginal} distributions of velocity at each
image point rather than the full probability distribution of the
entire velocity field, which would require spatial coupling.}. This
rejection of data allows the theory to implement basic temporal
grouping (for speculations about more complex temporal grouping, see
Yuille and Grzywacz 1998).  This basic form of temporal grouping is
related to data association as studied by engineers (Bar-Shalom and
Fortmann 1988) but differs by being probabilistic (following Ho and
Lee 1964). Finally, we show that our temporal grouping theory can be
implemented using locally connected networks, which are suggestive of
cortical structures.

The paper is organized as follows: The theoretical framework for
motion prediction and estimation is introduced in
Section~\ref{theory}. Section~\ref{model} gives a description of the
specific model chosen to implement the theory in the continuous time
domain. Implementation of the model using a locally connected network
is addressed in Section~\ref{implementation} and
Section~\ref{methods}. The model's predictions are tested in
Section~\ref{results} by performing computer simulations on the motion
phenomena described previously. Finally, we discuss relevant issues,
such as the possible relationship of the model to biology, in
Section~\ref{discussion}.

\section{The Theoretical Framework}
\label{theory}

\subsection{Background}

The theory of stochastic processes gives us a mathematical framework
for modelling motion signals that vary over time. These processes can
be used to predict the probabilities of future states of the system
(e.g. the future velocities) and are called Markov if the probability
of future states depends only on their present state (i.e.  {\it not}
on the time history of how they arrived at this state). In computer
vision, Markov processes have been used to model temporal coherence
for applications such as the optimal fusing of data from multiple
frames of measurements (Matthies {\it et al.} 1989; Clark and Yuille
1990; Chin {\it et al.}  1994). Such optimal fusing is typically
defined in terms of least-squares estimates which reduces to Kalman
filtering theory (Kalman 1960). Because Kalman filters are (recursive)
linear estimators that apply only to Gaussian densities, their
applicability in complex scenes involving several moving objects is
questionable. In this paper we propose a Bayesian formulation of
temporal coherence which is generalization of standard Kalman filters
and which can deal with targets moving in complex backgrounds.

We begin by deriving a generalization of Kalman filters which is
suitable for describing the temporal coherence of simple
motion\footnote{Our development follows the work of Ho and Lee 1964.}.
Consider the state vector $\vec{x}_k$ describing the status of a
system at time $t_k$ (in our theory $\vec x_k$ denotes the velocity
field at every point in the image). Our knowledge about how the system
might evolve to time $k+1$ is described in a probability distribution
function $p(\vec{x}_{k+1}|\vec{x}_k)$, known as the ``prior'' (in our
theory this prior will encode the temporal coherence assumption). The
measurement process that relates the observation $\vec z _k$ of the
state vector to its ``true'' state is described by the likelihood
distribution function $p(\vec{z}_k|\vec{x}_k)$ (in our theory $\vec
z_k$ will be the responses of basic motion units at every place in the
image). From these two distribution functions, it is straightforward
to derive the {\it a posteriori} distribution function
$p(\vec{x}_{k+1}|Z_{k+1})$, which is the distribution of $\vec{x}_{k+1}$
given the whole past set of measurements
$Z_{k+1}\doteq(\vec{z}_0,\cdots,\vec{z}_{k+1})$ (note that although
the prediction of the future depends only on the current state the
{\it estimate of the current state} does depend on the entire past
history of the measurements). Using Bayes' rule and a few algebraic
manipulations (Ho and Lee 1964), we get
\begin{equation}
\label{eq:estimation}
p(\vec{x}_{k+1}|Z_{k+1})=
\frac{p(\vec{z}_{k+1}|\vec{x}_{k+1})}{p(\vec{z}_{k+1}|Z_{k})}
p(\vec{x}_{k+1}|Z_{k}),
\end{equation}

 \noindent where $p(\vec{x}_{k+1}|Z_{k})$ is prediction for the future
state $\vec{x}_{k+1}$ given the current set of past measurements
$Z_{k}$, and $p(\vec{z}_{k+1}|Z_{k})$ is a normalizing term denoting
the confidence in the measure $\vec{z}_{k+1}$ given the set of past
measurements $Z_{k}$. The predictive distribution function can be
expressed as
\begin{equation}
\label{eq:prediction}
p(\vec{x}_{k+1}|Z_{k}) = \int
p(\vec{x}_{k+1}|\vec{x}_k)p(\vec{x}_{k}|Z_{k})d\vec{x}_k,
\end{equation}
 
Equations~\ref{eq:estimation},~\ref{eq:prediction} are generalizations
of the Wiener-Kalman solution for linear estimation in presence of
noise. Its evaluation involves two stages. In the {\it prediction
stage}, given by Equation~{\ref{eq:prediction}, the probability
distribution $p(\vec{x}_{k+1}|Z_{k})$ for the state at time $k+1$ is
determined. This function, which involves the present state and the
Markov transition describing the dynamics of the system, has a
variance larger than that of $p(\vec{x}_{k}|Z_{k})$. This increase in
the variance results from the non-deterministic nature of the motion.
In the {\it measurement stage}, performed by
Equation~\ref{eq:estimation}, the new measurements $\vec z_{k+1}$ are
combined using Bayes' theorem and, if consistent, reinforce the
prediction and decrease the uncertainty about the new
state. (Inconsistent measurements may increase the uncertainty still
further).

It was shown (Ho and Lee 1964) that if the measurement and prior
probabilities are both Gaussians, then Equations~\ref{eq:estimation}
and \ref{eq:prediction} reduce to the standard Kalman equations which
update the mean and the covariance of $p(\vec{x}_{k+1}|Z_{k+1})$ and
$p(\vec{x}_{k+1}|Z_{k})$ over time. Gaussian distributions, however,
are non-robust (Huber 1981) and an incorrect (outlier) measurement can
seriously distort the estimate of the true state. Standard linear
Kalman filter models are therefore {\it not able} to account for the
psychophysical data which demonstrate, for example, that human
observers are able to track targets despite the presence of
inconsistent (outlier) measurements (Watamaniuk {\it et al.}  1994,
Watamaniuk and McKee 1995). Various techniques, known collectively as
{\it data association} (Bar-Shalom and Fortmann 1988), can be applied
to reduce these distortions by using an additional stage which decides
whether to reject or accept the new measurements.  From the Bayesian
perspective, this extra stage is unnecessary and robustness can be
ensured by correct choices of the measurement and prior probability
models. More specifically, we specify a measurement model which is
robust against outliers.

\subsection{Prediction and Estimation Equations for a Flow Field}

We intend to estimate the motion flow field, defined over the
two-dimensional image array, at each time step. We describe the flow
field as $\{\vec v(\vec x,t)\}$, where the parentheses $\{\cdot\}$
denote variations over each possible spatial position $\vec x$ in the
image array.  The prior model for the motion field and the likelihood
for the velocity measurements are described by the probability
distribution functions $P(\{\vec v(\vec x,t)\}|\{\vec v(\vec
x,t-\delta)\})$ and $P(\{\vec\phi(\vec x,t)\}|\{\vec v(\vec x,t)\})$
(see Section~\ref{model} for details), where $\delta$ is time
increment and $\{\vec \phi(\vec x,t)\}$ represents the response of the
local motion measurement units over the image array. Our theory
requires that the local velocity measurements are normalized so that $
\sum _{\mu} \phi (\vec x,t) =1$ for all $\vec x$ and at all times $t$
(see Section~\ref{model} for details).  We let
$\Phi(t)=(\{\vec\phi(\vec x,t)\},\{\vec\phi(\vec
x,t-\delta)\},\cdots)$ be the set of all measurements up to and
including time $t$, and $P(\{\vec v(\vec
x,t-\delta)\}|\Phi(t-\delta))$ be the system's estimated probability
distribution of the velocity field at time $t-\delta$.  Using
Equations~\ref{eq:estimation} and \ref{eq:prediction} described above,
we get
\begin{equation}
P(\{\vec v(\vec x,t)\}|\Phi(t))=\frac{P(\{\vec\phi(\vec
x,t)\}|\{\vec v(\vec x,t)\})P(\{\vec v(\vec x,t)\}|\Phi(t-\delta))}
{P(\{\vec\phi(\vec x,t)\}|\Phi(t-\delta))},
\label{eq:motion-estimation}
\end{equation}

 \noindent for the estimation, and
\begin{eqnarray}
P(\{\vec v(\vec x,t)\}|\Phi(t-\delta)) =\mbox{\ \ \ \ \ \ \ \ \ \ \ \ \ \ \ \ \ \ \ \ \ \ \ \ \ \ \ \ \ \ \ \ \ \ \ \ \ \ \ \ \ \ \ \ \ \ \ \ \ \ \ \ \ \ \ \ \ \ }\nonumber\\
\int P(\{\vec v(\vec x,t)\}|\{\vec 
v(\vec x,t-\delta)\}) P(\{\vec v(\vec
x,t-\delta)\}|\Phi(t-\delta)) d[\{\vec v(\vec x,t-\delta)\}],
\label{eq:motion-prediction}
\end{eqnarray}
 
 \noindent for the prediction. 

\subsection{The Marginalization Approximation}

For simplicity and computational convenience, we assume that the prior
and the measurement distributions are chosen to be factorizable in
spatial position, so that the probabilities at one spatial point are
independent of those at another. This restriction prevents us from
including spatial coherence effects which are known to be important
aspects of motion perception (see, for example, Yuille and Grzywacz
1988). For the types of stimuli we are considering these spatial effects
are of little importance and can be ignored.

In mathematical terms, this spatial-factorization assumption means
that we can factorize the prior $P_p$ and the likelihood $P_l$ so that:
\begin{eqnarray} P_p(\{ \vec v (\vec x,t) \} | \{ \vec v(\vec x
^{\prime},t - \delta)\}) = \prod _{\vec x} p_p( \vec v (\vec x,t) | \{
\vec v(\vec x ^{\prime},t - \delta)\}) \nonumber \\ P_l(\{ \vec \phi
(\vec x,t)\}| \{\vec v (\vec x,t)\}) = \prod _{\vec x} p_l( \vec \phi
(\vec x,t)| \vec v (\vec x,t) \}).\end{eqnarray}

A further restriction is to modify
Equation~\ref{eq:motion-prediction} so that it updates the {\it
marginal distributions at each point independently}. This again
reduces spatial coherence because it decouples the estimates of
velocity at each point in space. Once again, we argue that this
approximation is valid for the class of stimuli we are
considering. This approximation will decrease computation time by
allowing us to update the predictions of the velocity distributions at
each point {\it independently}. The assumption that the measurement
model is spatially factorizable means that the estimation stage,
Equation~\ref{eq:motion-estimation}, can also be performed
independently.

This gives update rules for prediction $p_{pred}$:
 \begin{eqnarray} p_{pred}(\vec v (\vec x,t)| \Phi(t - \delta)) = \int [d
 \vec v (\vec x ^{\prime},t- \delta)] P_p(\vec v ^T(\vec x,t)|\{ \vec v
 (\vec x ^{\prime},t - \delta)\}) \nonumber \\ P_e(\{ \vec v (\vec x
 ^{\prime},t - \delta)\}| \Phi (t - \delta)).
 \label{eq:simplified-motion-prediction}\end{eqnarray}

 \noindent and for estimation $P_e$:
\begin{equation}
P_e(\vec v(\vec x,t)|\Phi(t))=\frac{P_l(\vec\phi(\vec
x,t)|\vec v(\vec x,t))P_{pred}(\vec v(\vec x,t)|\Phi(t-\delta))}
{P(\vec\phi(\vec x,t)|\Phi(t-\delta))}.
\label{eq:simplified-motion-estimation}
\end{equation}

In what follows we will consider a specific likelihood and prior
function which allows us to simplify these equations and define
probability distributions on the velocities at each point.  This will
lead to a formulation for motion prediction and estimation where
computation at each spatial position can be performed in parallel.

\section{The Model}
\label{model}

We now describe a specific model for the likelihood and prior
functions which, as we will show, can account for many psychophysical
experiments involving temporal motion coherence. Based on this
specific prior function, we then derive a formulation for motion
prediction in the continuous time domain.

\subsection{Likelihood Function}

The likelihood function gives a probabilistic interpretation of the
measurements given a specific motion. It is probabilistic because the
measurement is always corrupted by noise at the input stage. (In many
vision theories the likelihood function depends on the image formation
process and involves physical constraints such as, for instance, the
geometry and surface reflectance. Because we are considering
psychophysical stimuli, consisting of moving dots, we can ignore such
complications).

To determine our likelihood function, we must first specify the input
stage.  Ideally, this would involve modeling the behaviors of the
cortical cells sensing natural image motion, but this would be too
complex to be practical. Instead, we use a simplified model of a bank
of receptive fields tuned to various velocities
$\vec{v}_\mu(\vec{x},t)$, $\mu=1\cdots M$, and positioned at each
image pixel. These cells have observation activities $\{\phi _i(\vec
x,t)\}$ which are intended to represent the output of a neuronally
plausible motion model such as (Grzywacz and Yuille 1990).  (In our
implementation, these simple model cells receive contributions from
the motion of dots falling within a local neighbourhood, typically the
four nearest neighbours.  Intuitively, the closer the dot is to the
center of the receptive field, and the closer its velocity is to the
preferred velocity of the cell, then the larger the response.  The
spatial profile and velocity tuning curve of these receptive fields
are described by Gaussian functions whose covariance matrices
$\Sigma_{m:x}$ and $\Sigma_{m:v}$ depend on the direction of
$\vec{v}_\mu(\vec{x},t)$, and are specified in terms of their
longitudinal and transverse components $\sigma^2 _{m:x,L}$, $\sigma^2
_{m:x,T}$ and $\sigma^2 _{m:v,L}$, $\sigma^2 _{m:v,T}$. For more details,
see section 4.1 of (Yuille, Burgi, Grzywacz 1997)).

The likelihood function specifies the probability of the receptive
field responses conditioned on a ``true'' external motion field. We
assume that the measurements depend only on the velocity field at that
specific position. We can therefore write $P_l(\{\vec \phi(\vec x,t)\}|
\{ \vec v (\vec x,t)\}) = \prod _x P_l(\vec \phi(\vec x,t)| \vec v (\vec
x,t))$. We now specify:
\begin{equation}
\label{eq:likelihood}
P_l(\vec\phi(\vec{x},t)|\vec{v}_\mu(\vec{x},t))=
\frac{\Psi_l(\phi_1(\vec{x},t),\phi_2(\vec{x},t),\cdots,\phi_M(\vec{x},t),
\vec{v}_\mu(\vec{x},t))}{\int\cdots\int
P_J(\xi_1,\xi_2,\cdots,\xi_M,\vec{v}_\mu(\vec{x},t))d\xi_1d\xi_2\cdots d\xi_M},
\end{equation}

 \noindent where $P_J$ is the joint probability distribution and we
 set $\Psi _l$ to be\footnote{This is only one of several possible choices.
It is attractive, see (Yuille, Burgi, Grzywacz 1997) because it leads to
a simple linear update rule.}:
\begin{equation}
\label{eq:likelihood2}
\Psi_l(\phi_1(\vec{x},t),\phi_2(\vec{x},t),\cdots,\phi_M(\vec{x},t),
\vec{v}_\mu(\vec{x},t))=
\vec\phi(\vec{x},t)\cdot\vec{f}(\vec{v}_\mu(\vec{x},t)),
\end{equation}

\noindent with tuning curves $f$ given by:
\begin{equation}
\label{eq:f_i}
f_i(\vec{v}_\mu(\vec{x},t))=\frac{e^{-(1/2)(\vec{v}_i-\vec{v}_\mu)^T
\Sigma^{-1}(\vec{v}_i-\vec{v}_\mu)}}{\sum_{j=1}^M
e^{-(1/2)(\vec{v}_j-\vec{v}_\mu)^T
\Sigma^{-1}(\vec{v}_j-\vec{v}_\mu)}},
\end{equation}

 \noindent where $\Sigma$ is the covariance matrix which depends on
the direction of $\vec{v}_\mu(\vec{x},t)$, and is specified in terms
of its longitudinal and transverse components $\sigma^2 _{l:v,L}$ and
$\sigma^2 _{l:v,T}$).  The experimental data (Anstis and Ramachandran
1987, Gottsdanker 1956, Werkhoven {\it et al.} 1992) suggest that
temporal integration occurs for velocity direction rather than for
speed.  This is built into our model by choosing the covariances so
that the variance is bigger in the direction of motion than in the
perpendicular direction (i.e. the velocity component perpendicular to
the motion has mean zero and very small variance so the {\it
direction} of motion is fairly accurate, but the variance of the
velocity component along the direction of motion is bigger which means
that the estimation of the speed is not accurate). Observe that we
have assumed that the response of the measurement device is
instantaneous. It would be possible to adapt the likelihood function
to allow for a time lag but we have not pursued this. Such a model
might be needed to account for motion blurring.

\subsection{Prior Function} 
\label{prior}

We now turn to the prior function for position and velocity. For a dot
moving at approximately constant velocity, the state transitions for
position and velocity is given respectively by $\vec{x}(t+\delta) =
\vec{x}(t) + \delta\vec{v}(t) + \vec{\omega}_x(t)$, and
$\vec{v}(t+\delta) = \vec{v}(t) + \vec{\omega}_v(t)$, where
$\vec{\omega}_x(t)$ and $\vec{\omega}_v(t)$ are random variables
representing uncertainty in the position and velocity.  Extending this
model to apply to the flow of dots in an image requires a conditional
probability distribution $P_m(\vec v(\vec
x,t)|\{\vec{v}_\mu(\vec{x}_i,t-\delta)\})$, where the set of spatial
positions and the set of allowed velocities are discretized with the
spatial positions set to be $\{\vec{x}_i : i=1,\cdots,N\}$, and the
velocities at $\vec{x}_i$ to $\{\vec{v}_\mu : \mu=1,\cdots,M\}$. We
are assuming that $P_m(\{ \vec v(\vec x,t)\}|.) = \prod _x P_m( \vec
v(\vec x,t)|.)$ so that the velocities are predicted {\it independent}
of each other. We also assume that the prior is built out of two
components: (i) the probability
$p(\vec{x}|\vec{x}_i,\vec{v}_\mu(\vec{x}_i,t-\delta))$ that a dot at
position $\vec x_i$ with velocity $\vec v _{\mu}$ at time $t- \delta$
will be present at $\vec x$ at time $t$, and (ii) the probability
$P(\vec v(\vec x,t)|\vec{v}_\mu(\vec{x}_i,t-\delta))$ that it will have
velocity $\vec v(\vec x,t)$. These components are combined to give:

\begin{eqnarray}
\label{eq:multiple-dot-prediction}
P_m(\vec v(\vec
x,t)|\{\vec{v}_\mu(\vec{x}_i,t-\delta)\})=\mbox{\ \ \ \ \ \ \ \ \ \ \ \ \ \ \ \ \ \ \ \ \ \ \ \ \ \ \ \ \ \ \ \ \ \ \ \ \ \ \ \ \ \ \ \ \ \ \ \ \ \ \ \ }
\nonumber\\
\frac{1}{K(\vec{x},t)}\sum_i\sum_\mu P(\vec v(\vec
x,t)|\vec{v}_\mu(\vec{x}_i,t-\delta))\,p(\vec{x}|\vec{x}_i,\vec{v}_\mu(\vec{x}_i,t-\delta)),
\end{eqnarray}

 \noindent where $K(\vec{x},t)$ is a normalization factor chosen to
ensure that \linebreak $P_m(\vec v(\vec
x,t)|\{\vec{v}_\mu(\vec{x}_i,t-\delta)\})$ is normalized.  The two
conditional probability distribution functions in
Equation~\ref{eq:multiple-dot-prediction} are assumed to be normally
distributed, and thus,
\begin{eqnarray}
\label{eq:trajectory}
p(\vec{x}|\vec{x}_i,\vec{v}_\mu(\vec{x}_i,t-\delta)) \sim 
e^{-(1/2)(\vec{x}-\vec{x}_i-\delta\vec{v}_\mu(\vec{x}_i,t-\delta))^T
\Sigma^{-1}_x
(\vec{x}-\vec{x}_i-\delta\vec{v}_\mu(\vec{x}_i,t-\delta))}\\
\label{eq:constant-velocity}
P(\vec v(\vec x,t)|\vec{v}_\mu(\vec{x}_i,t-\delta)) \sim
e^{-(1/2)(\vec{v}(\vec{x},t)-\vec{v}_\mu(\vec{x}_i,t-\delta))^T
\Sigma^{-1}_v
(\vec{v}(\vec{x},t)-\vec{v}_\mu(\vec{x}_i,t-\delta))}.
\end{eqnarray}

 \noindent These functions express the belief that the dots are, on
average, moving along a straight trajectory
(Equation~\ref{eq:trajectory}) with constant velocity
(Equation~\ref{eq:constant-velocity}). The covariances $\Sigma_x$ and
$\Sigma_v$, which quantify the statistical deviations from the model,
are defined in a coordinate system based on the velocity vector
$\vec{v}_\mu$. These matrices are diagonal in this coordinate system and are
expressed in terms of their longitudinal and transverse components
$\sigma^2 _{x,L}, \sigma^2 _{x,T}, \sigma^2 _{v,L}, \sigma^2 _{v,T}$.

\subsection{Continuous Prediction}

Computer simulations of Equation~\ref{eq:simplified-motion-prediction}
(even for the simple case of moving dots) requires large-kernel
convolutions which require excessive computational time. Instead, we
re-express Equation~\ref{eq:simplified-motion-prediction} in the
continuous time domain. (Such a re-expression is not always possible
and depends on the specific choices of probability distributions. Note
that a similar approach has been applied for the computation of
stochastic completion fields by Williams and Jacobs 1997a,b). As shown
in the Appendix for Gaussian priors, the evolution of $P(\vec v(\vec
x,t)|\Phi(t-\delta))$ for $\delta\rightarrow 0$ satisfies a variant of
a partial differential equation, known as {\it Kolmogorov's forward
equation} or {\it Fokker-Planck} equation. Our equation is a
non-standard variant because, unlike standard Kolmogorov theory, our
theory involves probability distributions at every point in space
which interact over time. It is given by: \begin{eqnarray} {{\partial
}\over{\partial t}} P(\vec v (\vec x,t)) = {{1}\over{2}} \{ {{\partial
^T}\over{\partial \vec v}} {\bf \Sigma _{\vec v}} {{\partial
}\over{\partial \vec v}} \} P(\vec v(\vec x,t)) + {{1}\over{2}} \{
{{\partial ^T}\over{\partial \vec x}} {\bf \Sigma _{\vec x}}
{{\partial }\over{\partial \vec x}} \} P(\vec v(\vec x,t)) \nonumber
\\ - \vec v \cdot {{\partial}\over{\partial \vec x}} P(\vec v(\vec
x,t)) + {{1}\over{\left| V \right|}} {{\partial }\over{\partial \vec
x}} \int \vec v P(\vec v(\vec x,t)) d \vec v. \label{eq:neoKol}
\label{eq:kolmogorov}
\end{eqnarray}

 \noindent where $\left| V \right|$ is the volume of velocity space
$\int d \vec v$, and $\{ {{\partial ^T}\over{\partial \vec x}} {\bf
\Sigma_{\vec x}} {{\partial }\over{\partial \vec x}} \}$ and $\{
{{\partial ^T}\over{\partial \vec v}} {\bf \Sigma_{\vec v}} {{\partial
}\over{\partial \vec v}} \}$ are scalar differential operators (for
isotropic diffusions, these scalar operators are Laplacian
operators). The terms on the right hand side of this equation stand
for velocity diffusion (first term), spatial diffusion (second term)
with deterministic drift (third term), and normalization (fourth
term).  This equation is local and describes the spatio-temporal
evolution of the probability distributions $P(\vec v(\vec x,t))$.

\section{The Implementation}
\label{implementation}

We now address the implementation of the update rule for motion
estimation (Equations~\ref{eq:simplified-motion-estimation}). At each
spatial position, two separate cells' populations are assumed, one for
coding the likelihood function and another for coding motion
estimation. If we were to choose the large-kernel convolution for
motion prediction (Equation~\ref{eq:simplified-motion-prediction}),
then the network implementation would reduce to two positive linear
networks multiplying each other followed by a normalization,
illustrated in Figure~\ref{fig:network} and see (Yuille et al
1998). But a problem with implementing this network on serial
computers is that the range of the connections between
motion-estimation cells depends on the magnitude of $\vec{v}_\mu$
(that is, we would need long-range connections for high speed
tunings). Such long range connections occur in biological systems,
like the visual cortex, but are not well suited to VLSI or serial
computer implementations. Alternatively, using our differential
equation for predicting motion (Equation~\ref{eq:kolmogorov}) involves
only local connections. We now focus on this particular
implementation (see Figure~\ref{fig:network2}).

\begin{figure}
\centering
\includegraphics[height=4.5in]{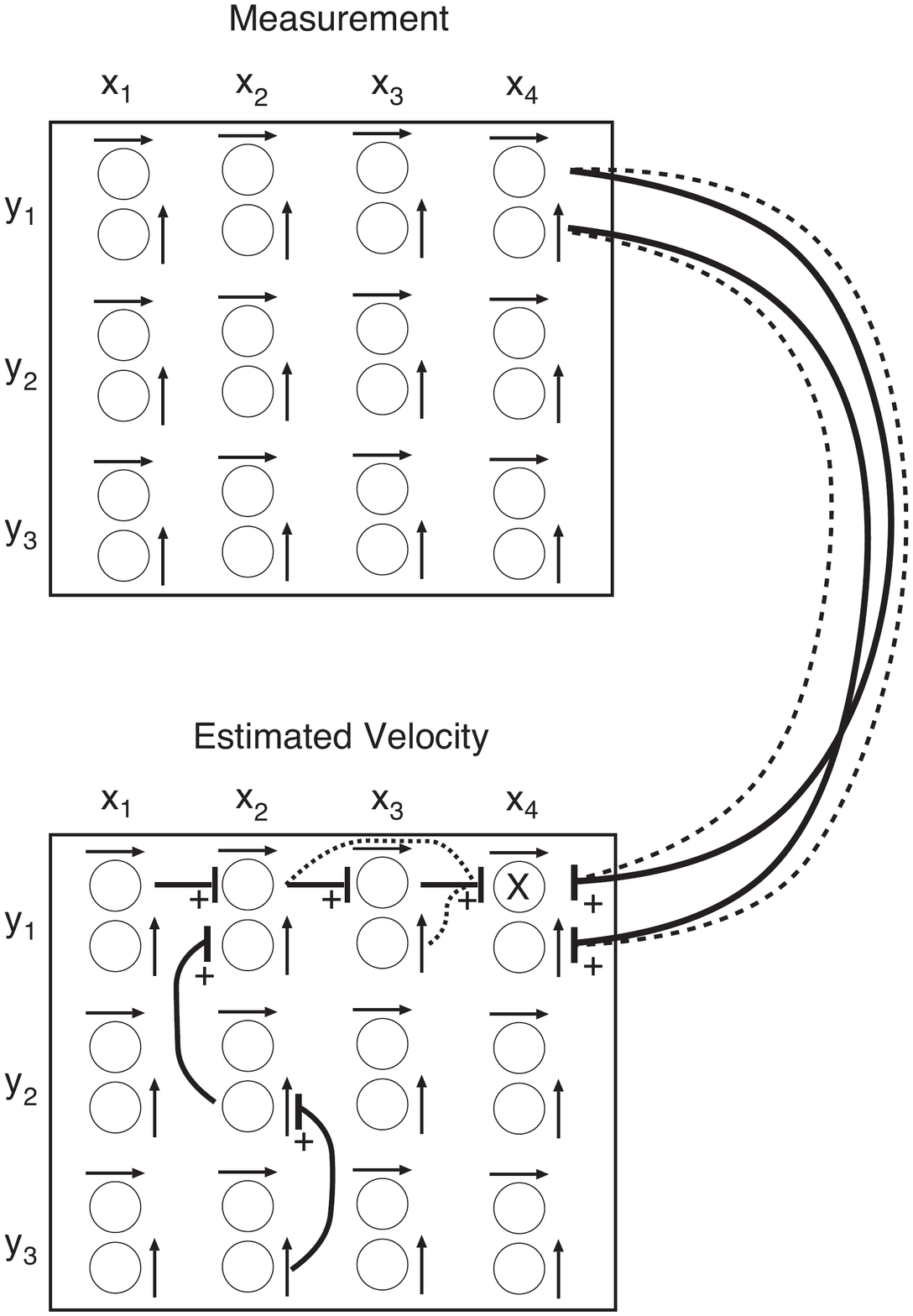}
\caption{Network for Motion Estimation. The network consists of two
    interacting layers. The top square shows the observation layer
    consisting of cells organized in columns on the $(x,y)$
    lattice. At each of the twelve spatial positions there is a
    velocity column which we display by two cells, shown as circles,
    with the adjacent arrows indicating their velocity tunings (here
    either horizontal or vertical). The lower square represents the
    estimation layer, which also consists of cells organized as
    columns on the $(x,y)$ lattice. In the measurement stage, the
    observation cells influence the estimation cells by excitatory
    (indicated by the ``plus'' sign) connections and multiplicative
    interactions (indicated by $\times$).  The excitation is higher
    between cells with similar velocity tuning, which we indicate by
    strong (solid lines) or weak (dashed lines) connections. In the
    prediction stage, cells within the estimation layer excite each
    other (again with the strength of the excitation being largest for
    cells with similar velocity tuning). Inhibitory connections within
    the columns are used to ensure normalization.}
\label{fig:network}
\end{figure}

\begin{figure}
\centering
\includegraphics[height=4.5in]{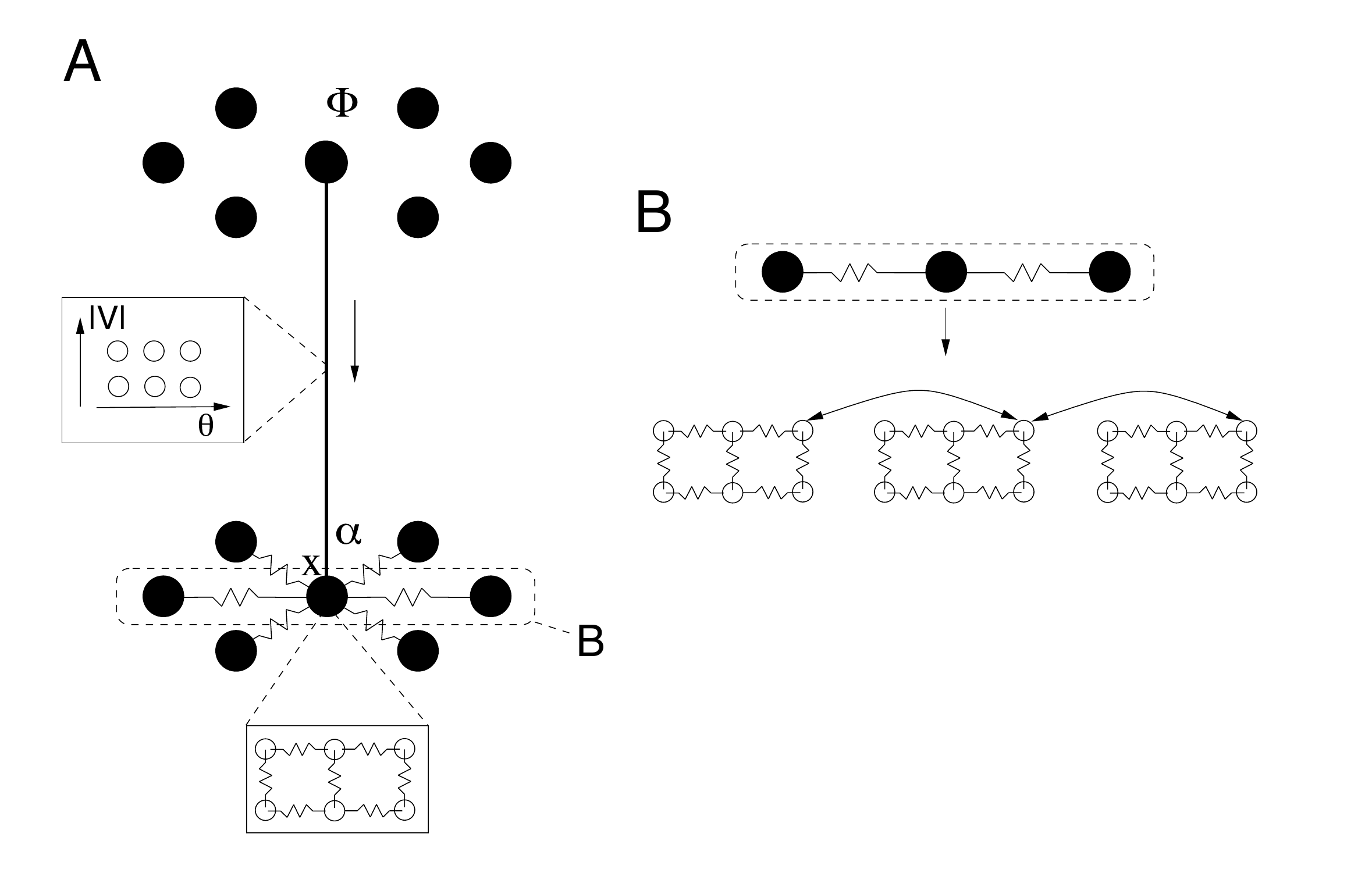}
\caption{Network for Motion Estimation Using Continuous
    Prediction. (A) Continuous motion prediction (lower part of the
    network) is accomplished through nearest neighbor spatial
    interactions, in this case according to a hexagonal lattice. The
    velocity distribution is represented at each spatial position by a
    set of velocity cells (small circles) organized according to a
    polar representation (small panel).  Observation cells in
    measurement layer (upper part) influence the estimation cells by
    multiplicative interactions (indicated by $x$) to yield motion
    estimation $\alpha$. (B) Interactions between two populations of
    velocity cells involves cells tuned to the same velocities, while
    interactions within a population involves cells tuned to different
    velocities.}
\label{fig:network2}
\end{figure}

Kolmogorov's equation can be solved on an arbitrary lattice using
a Taylor series expansion. The spatial term, which contains the
diffusion and drift terms, can be written so that the partial
derivatives at lattice points are functions of the values of the
closest neighbor points. For the sake of numerical accuracy, we have
chosen a spatial mesh system where the points are arranged
hexagonally. Furthermore, we found it convenient to express the
velocity vectors in polar coordinates where their norm $s$ and angle
$\theta$ can be represented on a rectangular mesh. The change of
variable is given by $\tilde{p}(s,\theta)= r p(\vec{v})$.  In polar
coordinates, the differential operator ${{\partial ^T}\over{\partial
\vec v}} {\bf \Sigma _{\vec v}} {{\partial }\over{\partial \vec v}}$
becomes
\begin{equation}
L_v[\tilde{p}] =
 \frac{1}{2}\sigma_{v,L}^2\frac{\partial^2\tilde{p}}{\partial r^2} -
 (\sigma_{v,L}^2-\frac{1}{2}\sigma_{v,T}^2)\frac{\partial}{\partial
 r}\left(\frac{1}{r}\,\tilde{p}\right)+
 \frac{1}{2}\sigma_{v,T}^2\frac{\partial^2\tilde{p}}{\partial
 \theta^2}\,\frac{1}{r^2},
\end{equation}

 \noindent where $\sigma^2_{v,L}$ and $\sigma^2_{v,T}$ are the
variances in the longitudinal and transverse directions, respectively.

We let the activities of the motion-estimation cells at time $t$ be
represented by a vector $\vec \alpha (\vec x,t) = (\alpha _1 (\vec
x,t),...,\alpha _M(\vec x,t))$. The iterative method for determining
such activities involves a four-steps scheme. The first three steps
are concerned with solving Kolmogorov's equation. For a cell at
position $\vec x=(x,y)$ and of velocity tuning $\vec
v_\mu=(s,\theta)$, the three steps are:

\begin{eqnarray}
\label{eq:pred2}
\alpha^{t+1/4}_{x,y,s,\theta}& =& \alpha^{t}_{x,y,s,\theta} + \Delta
t\sum_{\imath,\jmath}A_{\imath,\jmath}(s)\alpha^t_{x,y,s+\imath\Delta
s,\theta+\jmath\Delta\theta}\nonumber\\ \alpha^{t+1/2}_{x,y,s,\theta}&
=& \alpha^{t+1/4}_{x,y,s,\theta} + \Delta
t\sum_{\iota,\chi}B_{\iota,\chi}(s,\theta)
\alpha^t_{x+\iota,y+\chi,s,\theta}\nonumber\\
\alpha^{t+3/4}_{x,y,s,\theta}& =& \alpha^{t+1/2}_{x,y,s,\theta} +
\Delta t \sum_{s^\prime,\theta^\prime}
\sum_{\imath,\jmath}A_{\imath,\jmath}(s)\alpha^t_{x,y,s^\prime+\imath\Delta
s,\theta^\prime+\jmath\Delta\theta},
\end{eqnarray}

 \noindent where the coefficients $A_{\imath,\jmath}$ and
$B_{\iota,\chi}$ are functions of $r$, $\theta$, and the covariance
matrices. The constants $\Delta s$, $\Delta \theta$ represent the
quantization  factors for $s$ and $\theta$. The superscripts $t+1/4$,
$t+1/2$, $t+3/4$ indicate the order in which these three steps
proceed (i.e. a single time step for the complete system is broken down
into four substeps for implementational convenience).

The first step requires evaluating the spatial differential operator
and involves six neighbor points (on the hexagonal spatial
lattice). The second step calculates the velocity differential
operator and involves four neighbor cells. Periodic boundary
conditions are assumed in space and velocity. The third step 
performs the normalization. To guarantee stability of the whole
iterative method, the time step $\Delta t$ has been determined using
von Neumann analysis (see Courant and Hilbert 1953), which considers
the independent solutions, or eigenmodes, of finite-difference
equations. Typically, the stability conditions involve ratios between
the time step and the quantization scales of space and velocity.

If we are performing prediction without measurement, then the fourth
step --  the determination of the activity of the motion-estimation
cells -- simply reduces to
\begin{equation}
\alpha^{t+1}_{x,y,r,\theta}=\alpha^{t+3/4}_{x,y,r,\theta}.
\end{equation}
 
 \noindent If there are measurements, then we apply the
motion-estimation equation
(Equation~\ref{eq:simplified-motion-estimation}), which consists of
multiplying the motion prediction with the likelihood function (as defined
by Equations~\ref{eq:likelihood} and
\ref{eq:likelihood2}). Therefore the complete update rule is
\begin{equation}
\label{eq:est2}
\alpha^{t+1}_{x,y,r,\theta} = {{1}\over{N(\vec
x,t+1)}}\alpha^{t+3/4}_{x,y,r,\theta}\cdot \sum_j
\phi_j(\vec{x},t)\cdot f_j(\vec{v}_\mu(\vec{x},t)),
\end{equation}

 \noindent where $N(\vec x,t+1)$ is a normalization constant to impose
$\sum _{\mu =1}^M \alpha _{\mu}(\vec x, t+1) =1, \ \forall \vec x,t$
so that we can interpret these activities as the probabilities of the
true velocities.

\section{Methods}
\label{methods}

Our model has two extreme forms of behavior depending on how well the
input data agree with the model predictions. If the data are generally in
agreement with the modelÕs predictions, then it behaves like a standard
Kalman ?filter (i.e., with gaussian distributions) and integrates out the noise.
At the other extreme, if the data are extremely different from the prediction,
then the model treats the data as outliers or distractors and ignores them.
We are mainly concerned here with situations where the model rejects data.
The precise situation where noise and distractors start getting confused is
very interesting, but we do not address it here.

We ?first checked the ability of our model to integrate out weak noisy
stimuli for tracking a single dot moving in an empty background. For this
situation, temporal grouping (data association) is straightforward, and so standard Kalman ?filters can be used. To evaluate the model, we ran a set of
trials that plotted the relative estimation of two velocities as a function of
the number of jumps. The graph we obtained (see Figure~\ref{fig:speeddiscrimination}) is very similar
to those reported in the psychophysical literature (McKee {\it el al.} 1986).
Henceforth, we assume that the model is able to integrate out noisy input.
For the remaining experiments, we assumed that the local velocity
measurements were ÒnoisyÓ in the sense that the measurement units specified a probability of velocity measurements rather than a single velocity.
However, the Òprobability of velocitiesÓ itself was not noisy.

We next tested our model on three psychophysical motion phenomena.
First, we considered the improved accuracy of velocity estimation of a single
dot over time (Snowden and Braddick, 1989; Watamaniuk {\it el al.} 1989; Welch {\it et al.} 1997; Ascher, 1998). Then we examined
the predictions for the motion occlusion experiments (Watamaniuk
and McKee, 1995). Finally we investigated the motion outlier experiments
(Watamaniuk {\it et al.} 1994).

For all simulations we set the parameters as follows. All dots were
moving at a speed of 6 jumps per second (jps), where one jump
corresponds to the distance separating two neighbouring pixels in
the horizontal direction.  The longitudinal and transverse components of the
covariances matrices were $\sigma_{m:x,L}=0.8 \mbox{ jump}$,
$\sigma_{m:x,T}=0.4 \mbox{ jump}$, $\sigma_{m:v,L}=3.2 \mbox{ jps}$,
and $\sigma_{m:v,T}=2.6 \mbox{ jps}$ for the measurement,
$\sigma_{l:v,L}=2.2 \mbox{ jps}$, and $\sigma_{l:v,T}=1.1 \mbox{ jps}$
for the likelihood function, and $\sigma_{x,L}=0.6 \mbox{ jump}$,
$\sigma_{x,T}=0.3 \mbox{ jump}$, $\sigma_{v,L}=0.8 \mbox{ jps}$, and
$\sigma_{v,T}=0.4 \mbox{ jps}$ for the prior function. These
parameters were chosen by experimenting with the computer
implementation but the results are relatively insensitive to their
precise values (i.e. detailed fine tuning was not required).

Except for one of the occluder experiments (the occluder defined by
distractor motion), where we used 18 velocity channels (six
equidistant directions and three speeds), we used 30 velocity channels
positioned at each spatial position, that is, six equidistant
directions (thus $\Delta\theta=60\mbox{ deg}$, starting at the
origin), and five speeds ($\Delta r =2 \mbox{ jps}$, with slowest
speed channel tuned to 2 jps). To guarantee stability of the numerical
method we set the time increment to $\Delta t=0.6\mbox{ ms}$. Initial
conditions were uniformly distributed density functions. There were
$32 \time 32$ spatial positions.

\begin{figure}
\centering
\includegraphics[width=\textwidth]{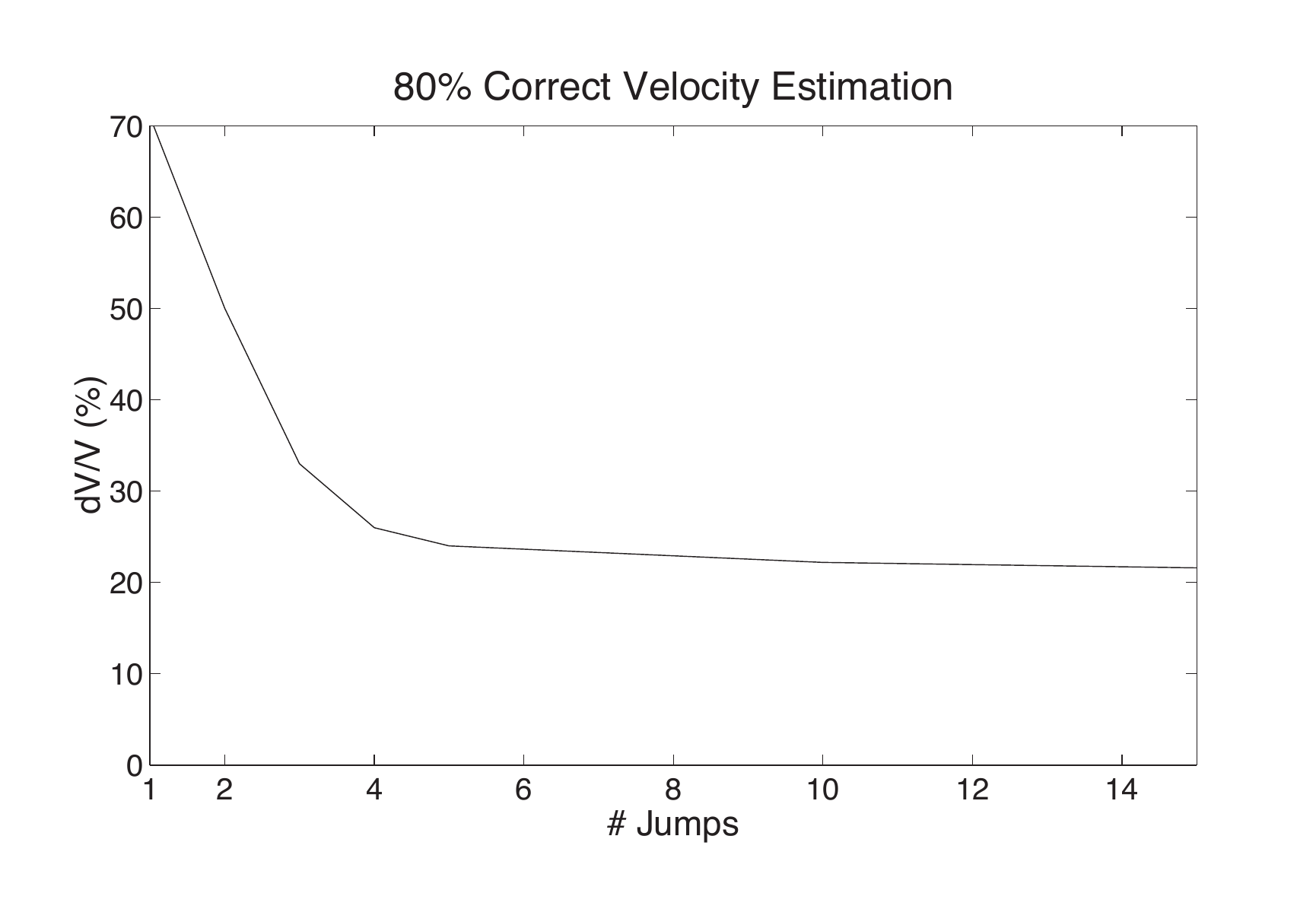}
\caption{Speed Discrimination For Two Moving Dots. This graph
    shows the improved ability to estimate the relative speed of two
    dots (one of which is moving at 2 jumps per second) as a function
    of the number of jumps. The level of 80\% correct discrimination
    is reached for diminishing speed differences (measured by $dV/V$)
    when the number of jumps increases, consistent with psychophysics
    (McKee {\it et al.}  1986).}
\label{fig:speeddiscrimination}
\end{figure}

\section{Results \label{results}}
\subsection{Velocity Estimation in Time}

For this experiment, the stimulus was a single dot moving in an empty
background. To evaluate how the velocity estimation evolves in time,
we measured two numbers: The first number was the {\it sharpness} of
the velocity probability at each position, which we define to be the
negative of the entropy of the distribution (the entropy of the
distribution is given by $- \sum _{\mu =1}^M \alpha _{\mu}(\vec x,t)
\log \alpha _{\mu}(\vec x,t)$) plus the maximum of the entropy (to
ensure that the sharpness is non-negative). Observe that the sharpness
is the {\it Kullback-Leibler divergence} $D(\alpha||U)$ between the
velocity distribution $\alpha _{\mu}(\vec x,t)$ and the uniform
distribution $U_{\mu} = 1/M, \ \forall \ \mu$ (more precisely
$D(\alpha||U)= \sum _{\mu =1}^M \alpha _{\mu}(\vec x,t) \log \{ \alpha
_{\mu}(\vec x,t) / U_{\mu} \}$.)  As is well-known (Cover and Thomas
1991) the Kullback-Leibler divergence is positive semi-definite taking
its minimum value of $0$ when $\alpha _{\mu} = U_{\mu}, \ \forall \mu$
and increases the more $\alpha _{\mu}$ differs from the uniform distribution
$U_{\mu}$ (i.e. the {\it sharper} the distribution $\alpha _{\mu}$
becomes). Hence the higher this sharpness, the more precise the velocity
estimate. Moreover, the sharpness will be lowest in positions where
there are no dots moving (i.e. for which $\phi_\mu(\vec x,t) =1/M$,
$\mu=1,\cdots, M$).  The target, a coherently moving dot, should have
a relatively high sharpness response surrounded by low sharpness
responses in neighboring positions.  The second number is the {\it
confidence factor} $P( \vec \phi (\vec x,t)| \Phi(t- \delta))$ in
Equation~\ref{eq:simplified-motion-estimation}. This measures how
probable the model judges the new input data. The higher this number,
the more the new measurement is consistent with the prediction (and
hence the more likely that the measurement is due to coherent motion).

The results of motion estimation for this experimental stimulus is
shown in Figure~\ref{fig:velest}.  This figure shows an initial fast
decrease in the entropy of the velocity distribution followed by a
slow decrease (i.e. an increase in sharpness of the density
function). Also shown is the increase in the confidence factor of the
target velocity estimation at each new iteration. These two effects
indicate the increased accuracy of the velocity estimation with each
new iteration, consistent with the psychophysical literature (Snowdon
and Braddick 1989, Watamaniuk {\it el al.} 1989, Welch {\it et al.}
1997, Ascher 1998).  Note that for coherent motion the sharpness of
the model increases rapidly at first and then appears to slow
down. Our results suggest that the model reaches an asymptotic limit
although they do not completely rule out a continual gradual
increase. We argue that asymptotic behaviour is more likely as both
the predictions and the measurements have inherent noise which can
never be eliminated. Moreover, it can be proven that the standard
(Gaussian) Kalman model will converge to an asymptotic limit depending
of the variances of the prior and likelihood functions (and we
verified this by computer simulations).  It seems also that human
observers also reach a similar asymptotic limit and their accuracy
does not become infinitely good with increasing number of input
stimuli.

We also tested the estimation of velocity for a dot moving on a
circular trajectory and our model can successfully estimate the dot's
speed, although the sharpness and confidence are not as high as they
are for straight motion (recall that the prior prefers straight
trajectories). This again seems consistent with the psychophysics
(Watamaniuk {\it et al.} 1994, Verghese {\it et al.} 1999).

\begin{figure}
\centering
\includegraphics[width=0.8\textwidth]{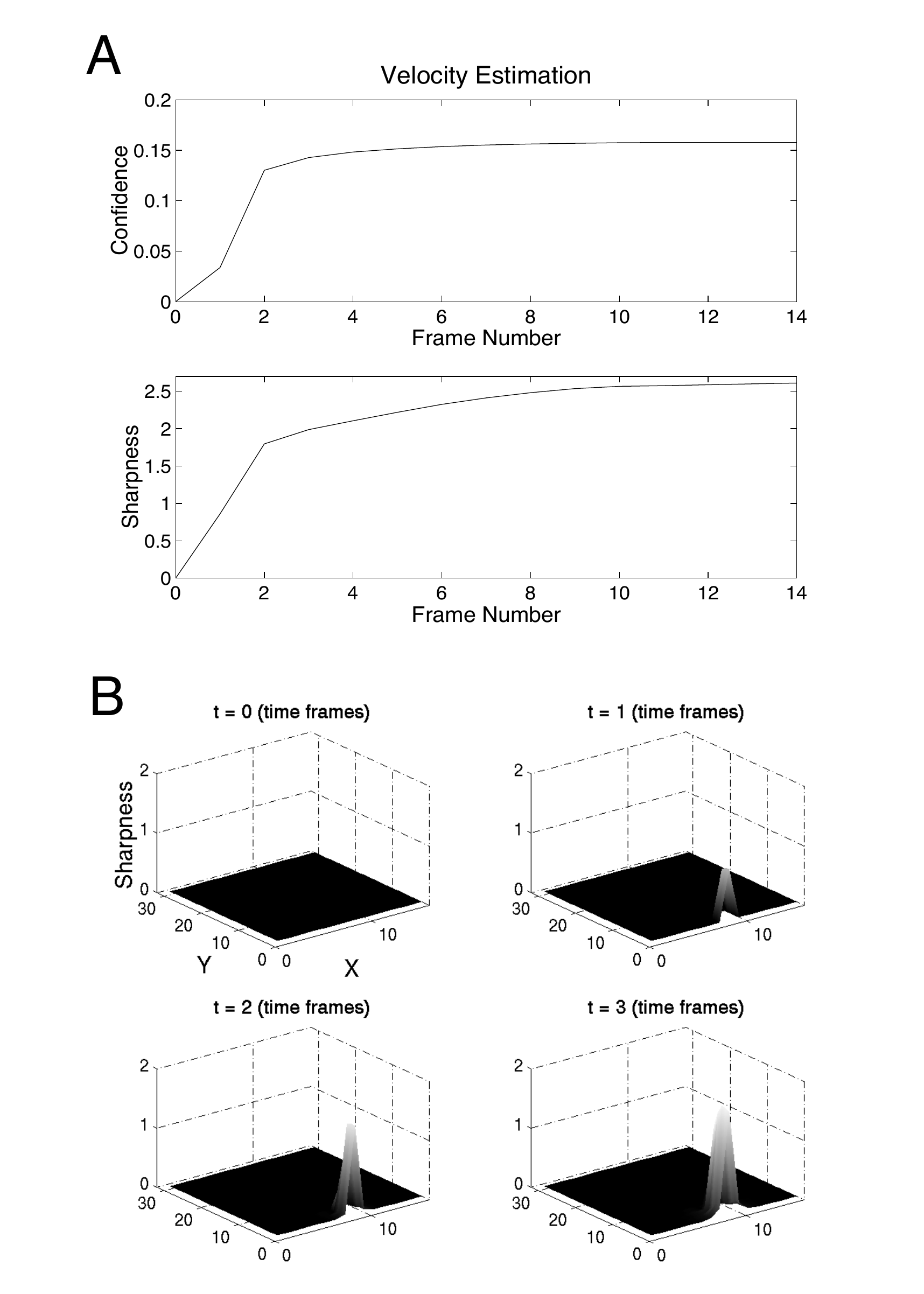}
\caption{Single Dot Experiment. A single dot is
    moving with constant velocity. Increasing accuracy of the velocity
    estimate with time is visible by an increase in the confidence
    factor and in the sharpness of the density function. Observe how
    the both confidence and sharpness increase rapidly at first but
    then quickly start to flatten out. (Our model outputs a datapoint
at each time frame and our plotting package joins them together
by straight-line sgements.)}
\label{fig:velest}
\end{figure}

\subsection{Single Dot With Occluders}

Next, we explored what would happen if the target dot went behind an
occluding region where no input measurements could be made (we call
this a {\it black occluder}). The results were the same as the previous
case until the target dot reached the occluder. In the occluded
region, the motion model continued to propagate the target dot but,
lacking supporting observations, the probability distribution started
to diffuse in space and in velocity. This dual diffusion can be seen
in Figure~\ref{fig:occluder} where the sharpness of velocity
estimation is shown after the dot entered the occluding
region. Observe the decrease in sharpness of the density function,
indicating a degradation of velocity estimation, when the target is
behind the occluder. However, the model still has enough ``momentum''
to propagate its estimate of the target dot's position even though no
measurements are available (this will, of course, breakdown if the
occluder gets too large). This is consistent with the findings by
Watamaniuk and McKee (1995) who showed that observers had a higher
than average chance of detecting the target when it emerges from the
occluder into a noisy background.

\begin{figure}
\centering
\includegraphics[width=0.8\textwidth]{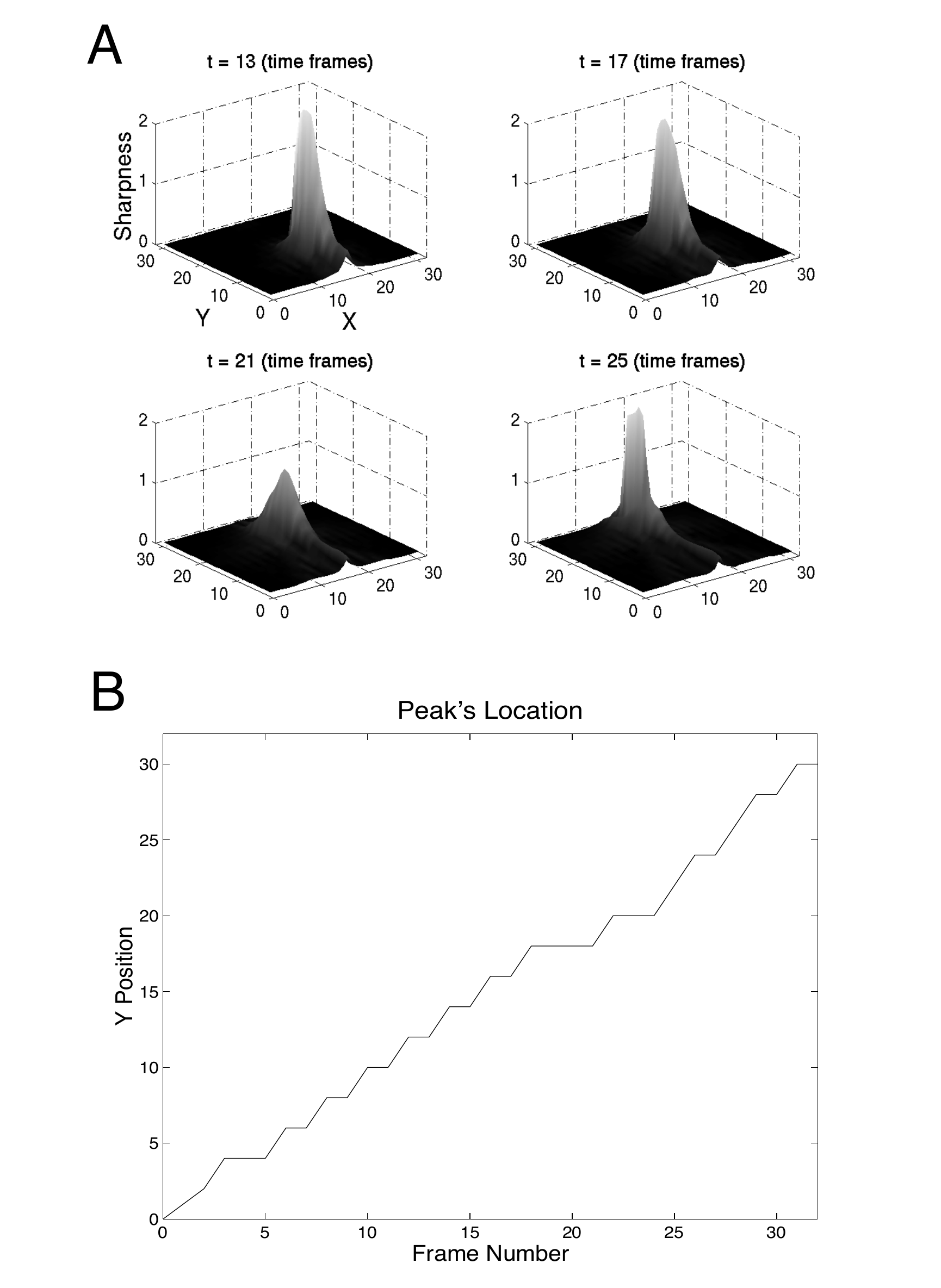}
\caption{Single Dot Experiment With Black
    Occluder. (A) This Representation shows how the spatial blur and
    sharpness of the velocity distribution change with time (measured
    in time frames) when a single dot moving along the Y-axis gets
    occluded. The occluding area is from $Y=15$ up to $Y=22$.  (B) The
    Y-location of the peak of velocity distribution as a function of
    time. Motion inertia (momentum) keeps the peak moving during the
    occlusion, albeit with a tendency to slow down (as visible by the
    wider plateau around time frame 20).}
\label{fig:occluder}
\end{figure}

We then tested our model with {\it motion occluders}, occluders
defined by motion flows as described in Watamaniuk and McKee (1995),
see Figure~\ref{fig:occluder2}A. We use the same measures as for the
single isolated dot. We also plot the cells' activities as a function
of the direction tuning for the cells tuned to the optimal speed
(Figure~\ref{fig:occluder2}B). The plot indicates that the cells can
signal non-zero probabilities for several motions at the same time.
After entering the occluding region, the peak corresponding to the
target motion gets smaller than the peak due to the
occluders. However, the peak of the target remains and so the target
peak can rapidly increase as the target exits from the occluders. Our
model predicts that is easier to deal with black occluders than with
motion occluders which is not consistent with the psychophysics
that shows a rough similarity of effects for both types of occluders. We
offer two possible explanations for this inconsistency:  First, the
model's directionally selective units and/or prediction-propagation
mechanism have too broad a directional tuning. Narrowing it may lead to
weaker interactions between the motion occluder and the target.  Second,
in contrast with the black occluders, the motion occluders may group
together as a surface, which would help explain why the visual system may
be more tolerant of motion occluders.  (Such a heightened tolerance might
compensate for the initial putative advantage of the black occluders.)
Kanizsa (1979) demonstrates several perceptual phenomena, which appear to
require this type of explanation. A limitation of our current model is
that it does not include such effects.

\begin{figure}
\centering
\includegraphics[width=0.6\textwidth]{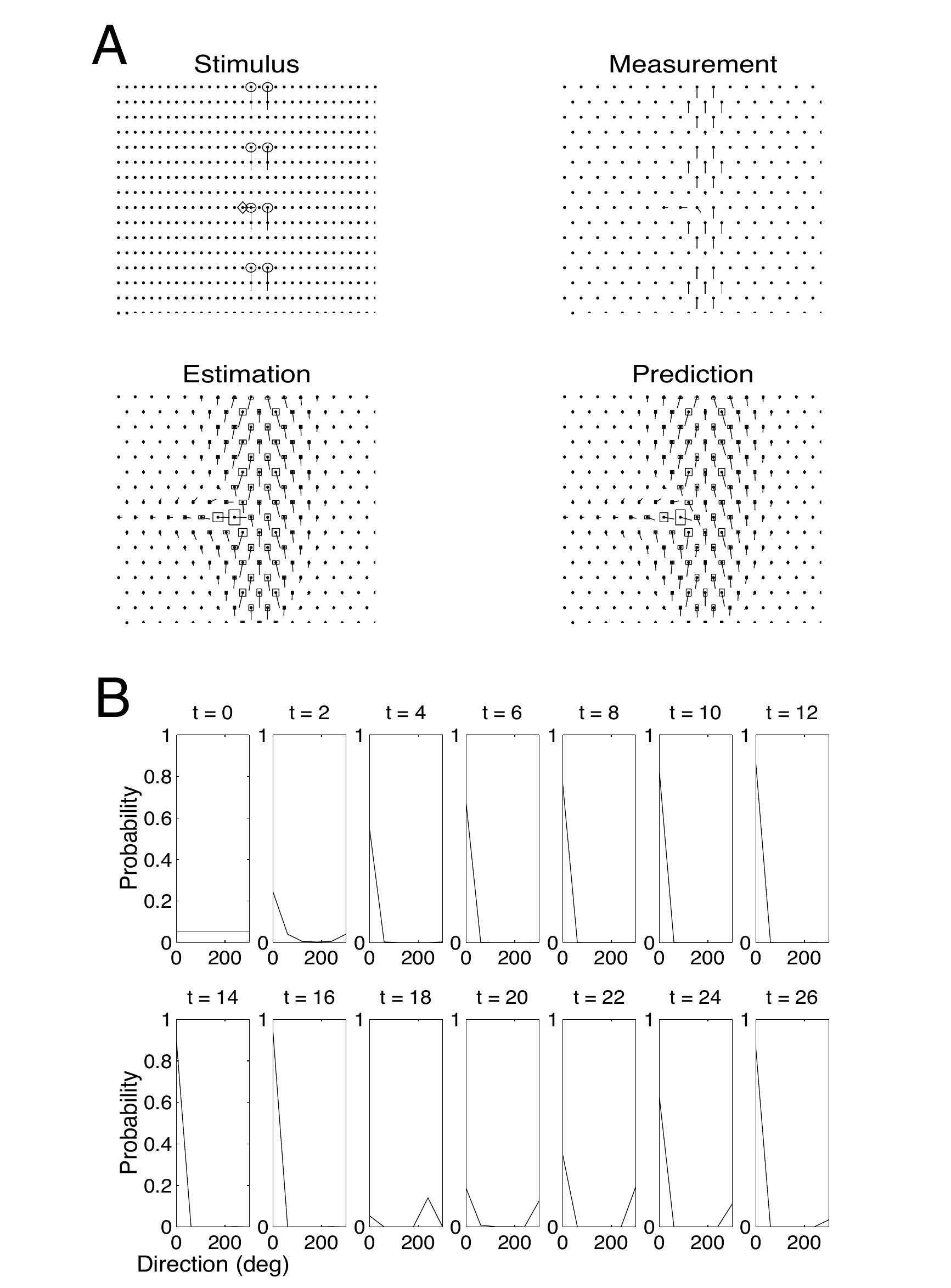}
\caption{Single Dot Experiment With
    Occluders Defined by Distractor Motion. (A) The occluder is
    defined by vertical motion of distractor dots, visible as circles
    in the left-top frame with arrows indicating velocity (the dots's
    speed is 6 jumps per second, and is represented by the length of
    the lines).  The target motion, not shown because it is hidden by
    the occluder, is moving horizontally from left to right. The
    motion-inertia effect of the target motion on the distractors is
    shown in the estimation and prediction stages where the height and
    width of the rectangles at each point indicate the confidence and
    sharpness, respectively (a small rectangle indicates low
    confidence and sharpness).  The lattice, marked with small dots,
    is hexagonal for the measurement, estimation, and prediction
    stages.  (B) The probability distribution of the velocity cells
    tuned to 6 jumps per second plotted as a function of directional
    tuning at different time steps.  As the target dot enters the
    occluder, at time frame 10, two peaks start developing in the
    probability distribution.  The bigger the occluder the more the
    peak induced by the motion of the distractor dots starts
    dominating. But as the dot re-emerges from the occluder it rapidly
    becomes sharp again, time frame 12.}
\label{fig:occluder2}
\end{figure}

\subsection{Outlier Detection}

In the outlier detection experiments (Watamaniuk {\it et al.} 1994,
Verghese {\it et al.}  1999), the target dot is
undergoing regular motion but it is surrounded by distractor dots,
which are undergoing random Brownian motion. To show the velocity
estimation at each position, we plot the response of our network using
a rectangle to display the two properties of sharpness and
confidence. The width of the rectangle gives the sharpness of the set
of cells at that point and the height gives the confidence factor. The
arrow shows the mean of the estimated velocity. It can be seen
(Figure~\ref{fig:outlier}) that the target dot's signal rapidly
reaches large sharpness and confidence by comparison to the distractor
dots, which are not moving coherently enough to gain confidence or
sharpness. The sharpness of the target dot's signal does not grow
monotonically because the distractor dots sometimes interfere with the
target's motion by causing distracting activity in the measurement
cells.

\begin{figure}
\centering
\includegraphics[width=\textwidth]{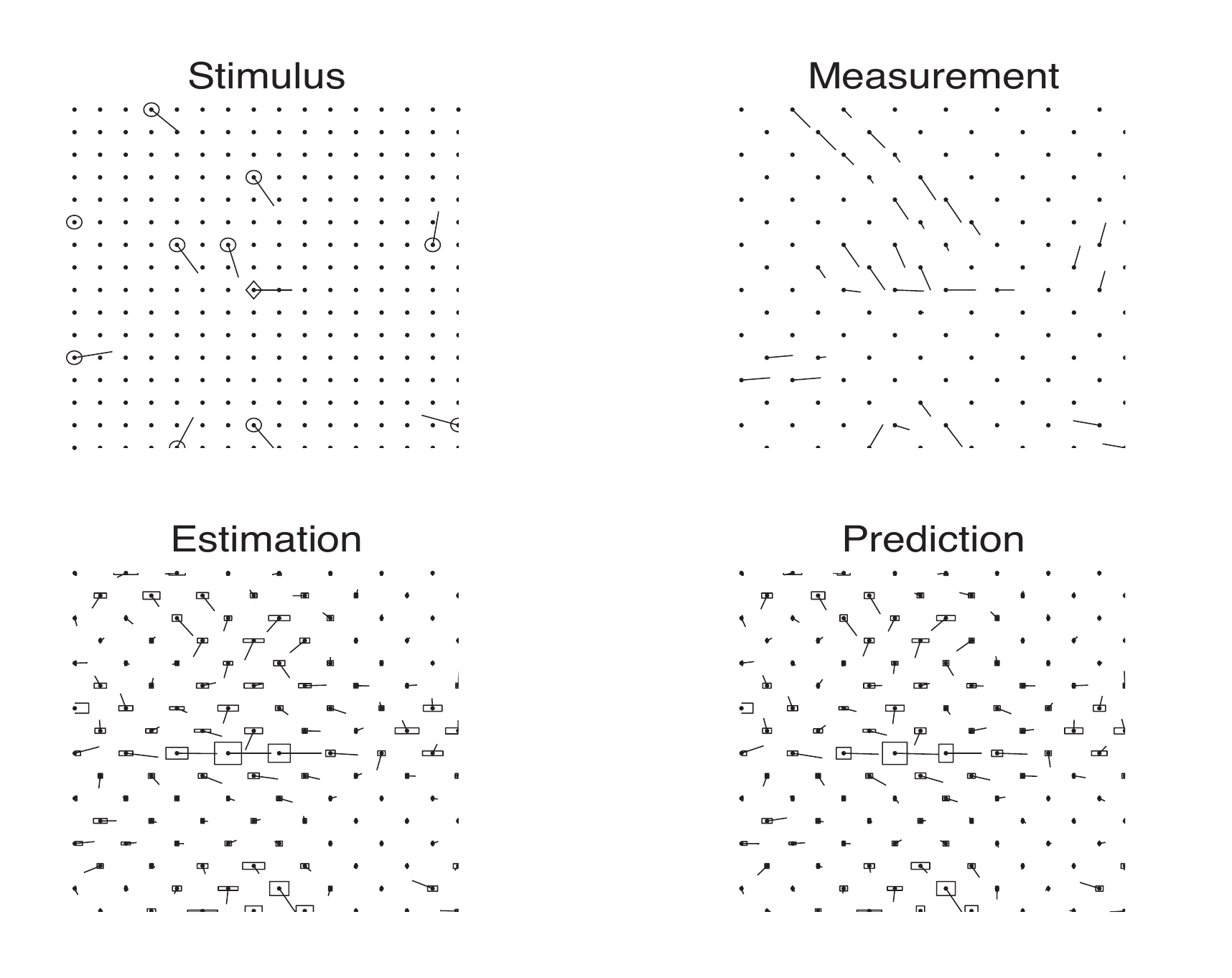}
\caption{Outlier Detection. A target dot
    (shown as a circle with line pointing rightwards indicating its
    velocity) is half way down the stimulus frame moving from left to
    right. It is surrounded by distractor dots undergoing random
    Brownian motion. After a few time steps the model gives high
    confidence and sharpness for the target dot. Observe that motion
    estimates of the distractor dots can sometimes become sharp if
    their motion direction is not changing too radically between two
    time frames (see the large rectangles at the bottom of the
    estimation and prediction panels. Graphic conventions as in
    previous figure.}
\label{fig:outlier}
\end{figure}

\section{Discussion}
\label{discussion}

This work provides a theory for temporal coherence. The theory was
formulated in terms of Bayesian estimation using motion measurements
and motion prediction. By incorporating robustness in the measurement
model, the system could perform temporal grouping (or data
association), which enabled it to choose what data should be used to
update the state estimation and what data could be ignored. We also
derived a continuous form for the prediction equation, and showed that
it corresponds to a variant of Kolmogorov's equations at each node. In
deriving our theory, we made an assumption of spatial factorizability
of the probability distributions and we made the approximation of
updating the marginal distributions of velocity at each point. This
allowed us to perform local computations and simplified our
implementation. We argued that these assumptions/approximations (as
embodied in our choice of prior probability) are suitable for the
stimuli we are considering, but that would need to be revised to
include spatial coherence effects (Yuille and Grzywacz 1988, Yuille
and Grzywacz 1989). The prior distribution used in this paper would
need to be modified to incorporate these effects.

In addition, recent results by McKee and Verghese (personal
communication) also suggest that that a more sophisticated prior may
be needed. Their results seem to show that observers are better at
predicting {\it where} there is likely to be motion rather than {\it
what} the velocity will be. Recall that our prior probability model
has two predictive components based on the current estimation of
velocity at each image position: First, the velocity estimation is
used to predict the positions in the image where the velocity cells
would be excited.  Second, we predict that the new velocity is close to
the original velocity. By contrast, McKee and Verghese's results seem
to show that human observers can cope with reversals of motion
direction (such as a ball bouncing off a wall). This is a topic of
current research and we note that priors which allow for this reversal
have already been developed by the computer vision tracking community
(Isard and Blake 1998).

Simulations of the model seem in qualitative agreement with a number
of existing psychophysical experiments on motion estimation over time,
motion occluders and motion outliers (McKee {\it et al.} 1986;
Watamaniuk and McKee 1995; Watamaniuk {\it et al.} 1994). Such
a qualitative agreement is characterized by three main features: (i) 
the final covariance of the motion estimate, (ii) the number of
jumps needed to reach such a final covariance, and (iii) the extent of
motion inertia during an occlusion, or between two measurements. These
three features are controlled as follows: the likelihood's covariance matrix
determines the initial distribution, and thus the number of jumps to
reach the desired covariance, while the prior's covariance matrix
affects all three features by imposing an upper bound to the
covariance and by setting the time constants of the diffusion process
between measurements.

Implementation of the model's equations led to a network that shares
interesting similarities with known properties of the visual
cortex. It involves a columnar structure where the columns, at each
spatial point, are composed of a set of cells tuned to a range of
velocities (columnar structures exist in the cortex but none have yet
been found for velocities). The observation-cell layer involves local
connections within the columns to compute the likelihood function. The
estimated velocity flow is computed in a second layer. If these layers
exist, it would be natural for them to lie in cortical areas involved
with motion such as MT and MST (Maunsell and Newsome 1987, Merigan and
Maunsell 1993). The estimation layer carries out calculation according
to Kolmogorov's equation, or the discrete long range version described
in (Yuille et al 1998), and consists of spatial columns of cells which
excite locally each other to predict the expected velocities in the
future. The calculation requires excitatory and inhibitory synapses
(i.e. no ``negative synapses''). The inputs from the observation layer
multiply the predictions in the estimated layer. Finally, we postulate
that inhibition occurs in each column to ensure normalization of the
velocity estimates at each column. This normalization could be
implemented by lateral inhibition as proposed by Heeger (1992).

A difficult aspect of this network, as a biological model, is its need
for multiplications (such multiplications arise naturally from our
probabilistic formulation using the rules for combining
probabilities).  Neuronal multiplication mechanisms were argued for by
Reichardt (1961), Poggio and Reichardt (1973, 1976) on theoretical
grounds. A specific biophysical model to approximate multiplication
was proposed by Torre and Poggio (1978). Detailed investigations of
this model, however, showed that it provided at best a rough
approximation for multiplication (Grzywacz and Koch 1987). Moreover,
experiments showed that motion computations in the rabbit retina, for
which the model was originally proposed, were not well approximated by
multiplications (Grzywacz {\it et al.} 1990).  Recent related work
(Mel {\it et al.} 1988, Mel 1993), though arguing for
multiplication-like processing, has not attempted to attain good
approximations to multiplication. On the other hand, the complexities
of neurons make it hard to rule out the existence of domains in which
they perform multiplications. Overall, it seems best to be agnostic
about neuronal multiplication and trust that more detailed experiments
will resolve this issue. More pragmatically, it may be best to favour
neuronal models which correspond to known biophysical mechanisms. The
biophysics of neurons may prevent them from performing ``perfect''
Bayesian computations and perhaps one should instead seek optimal
models which respect these biophysical limitations.

Finally, we emphasize that simple networks of the type we propose can
implement Bayesian generalizations of Kalman filters for estimation
and prediction. The popularity of standard linear Kalman filters is
often due to pragmatic reasons of computational efficiency. Thus
linear Kalman filters are often used even when they are inappropriate
as models of the underlying processes.  In contrast, the main drawback
of such Bayesian generalizations is the high computational cost, even
though statistical sampling methods can be used successfully in some
cases (see Isard and Blake 1996). Our study shows that these
computational costs may be better dealt with by using networks with
local connections and we are investigating the possibility of
implementing our model on parallel architectures in the expectation
that we will then be able to do Bayesian estimation of motion in real
time.

\newpage

\section*{Appendix}

\renewcommand{\theequation}{A\arabic{equation}}
\setcounter{equation}{0}

In this Appendix, we derive Kolmogorov's equation for motion
prediction. Let consider
Equation~\ref{eq:simplified-motion-prediction} for $\delta\rightarrow
0$: Taking this limit is tricky and requires It\^o calculus (Jazwinski
1970).  Our case is also non-standard because our theory involves
probability distributions at every point in space, which interact over
time.  This means that we cannot simply use the standard Kolmogorov's
equations, which apply to a single distribution only. Here we give a
simple derivation, however, which uses delta functions and is
appropriate when the prior model is based on a Gaussian $G(.)$. The
prior specified by equations \ref{eq:multiple-dot-prediction},
\ref{eq:trajectory} is written as:

\begin{eqnarray} P(\vec v (\vec x,t+ \delta)| \{ \vec v ^{\prime}
(\vec x ^{\prime},t) \}) = \int d \vec x ^{\prime} \int d \vec v
^{\prime}(\vec x ^{\prime}) G(\vec v (\vec x,t+ \delta) - \vec v
^{\prime} (\vec x ^{\prime},t): \delta {\bf \Sigma _{\vec v}})
\nonumber \\ \times {{ G(\vec x - \vec x ^{\prime} - \delta \vec v
^{\prime}(\vec x ^{\prime},t ): \delta {\bf \Sigma _{\vec x}})} \over
\int d \vec x ^{\prime \prime} \int d \vec v ^{\prime \prime} (\vec x
^{\prime \prime}) G(\vec x - \vec x ^{\prime \prime } - \delta \vec v
^{\prime \prime }(\vec x ^{\prime \prime},t); \delta {\bf \Sigma _{\vec x}})}. 
\label{eq:prior}
  \end{eqnarray}

\noindent The normalization term in the denominator is used to ensure
that the probabilities of the velocities integrate to one at each
spatial point $\vec x$.  In this equation, we have scaled the covariances
${\bf \Sigma}$ by $\delta$. This is necessary for taking the limit as
$\delta \mapsto 0$ (Jazwinski 1970).

Between observations we can express the evolution of the prior density
$P(\vec v (\vec x,t))$ as:

\begin{eqnarray} {{\partial P(\vec
v (\vec x,t))}\over{\partial t}} = \lim _{\delta \mapsto 0} \{
{{P(\vec v(\vec x,t + \delta) | \{ \vec v ^{\prime}(\vec x
^{\prime},t)\}) - P(\vec v (\vec x,t))} \over{\delta}} \nonumber \} \\
= \lim _{\delta \mapsto 0} {{1}\over{\delta}} \{ \int d \vec x
^{\prime} \int d \vec v ^{\prime} (\vec x ^{\prime}) P(\vec v (\vec
x,t + \delta)| \{ \vec v ^{\prime}(\vec x ^{\prime},t)\}) P(\{ \vec v
^{\prime}(\vec x ^{\prime},t)\}) - P(\vec v(\vec x,t)) \}
\label{eq:limit} \end{eqnarray}

 We now perform a Taylor series expansion of $P(\vec v(\vec x,t+
\delta) | \{ \vec v ^{\prime}(\vec x ^{\prime},t) \})$ in powers of
$\delta$ keeping the zeroth and first order terms (higher order terms
will vanish when we take the limit as $\delta \mapsto 0$). To perform
this expansion, we use the assumption that this distribution is
expressed in terms of Gaussians. As $\delta \mapsto 0$ these Gaussians
will tend towards delta functions and the derivatives of Gaussians
will tend towards derivatives of delta functions (thereby simplifying
the expansion). This derivation can be justified rigourously, playing
detailed attention to convergence issues, by the use of distribution
theory or the application of It\^o calculus.  If we expand $G(\vec v
(\vec x,t + \delta) - \vec v ^{\prime} (\vec x ^{\prime},t): \delta
{\bf \Sigma _{\vec v}})$ about $ \delta =0$ we find that the zeroth
order term is a Dirac delta function with argument $ \vec v (\vec x,t
+ \delta) - \vec v ^{\prime} (\vec x ^{\prime},t)$. This term can
therefore be integrated out. The derivative with respect to $\delta$
will effectively correspond to differentiating the Gaussian with
respect to the covariance. By standard properties of the Gaussian this
will be equivalent to a second-order spatial derivative. We will get
similar terms if we expand $ G(\vec x - \vec x ^{\prime} - \delta \vec
v ^{\prime}(\vec x ^{\prime},t ): \delta {\bf \Sigma _{\vec x}})$, but
we will also get an additional drift term from differentiating the
$\delta \vec v^{\prime}(\vec x ^{\prime},t)$ argument. In addition, we
will get other terms from the denominator of
Equation~\ref{eq:prior}. (These are required to normalize the
distributions and are non-standard. They are needed because we have a
differential equation for a set of interacting probability
distributions while the standard Kolmogorov's equation is for a single
probability distribution.)  We collect all these zeroth and first
order terms in the expansion and substitute into
Equation~\ref{eq:limit}. We can then evaluate the integrals using
known properties of the delta functions. After some algebra, these
integrals yield our variant of Kolmogorov's equation:

 \begin{eqnarray} {{\partial }\over{\partial t}} P(\vec v (\vec x,t))
 = {{1}\over{2}} \{ {{\partial ^T}\over{\partial \vec v}} {\bf \Sigma
 _{\vec v}} {{\partial }\over{\partial \vec v}} \} P(\vec v(\vec x,t))
+  {{1}\over{2}} \{ {{\partial ^T}\over{\partial \vec x}} {\bf \Sigma
 _{\vec x}} {{\partial }\over{\partial \vec x}} \} P(\vec v(\vec x,t))
 \nonumber \\
 - \vec v \cdot {{\partial}\over{\partial \vec x}} P(\vec v(\vec x,t))
 + {{1}\over{\left| V \right|}} {{\partial }\over{\partial \vec x}}
 \int \vec v P(\vec v(\vec x,t)) d \vec v.
\label{eqa:kolmogorov}
\end{eqnarray}

 \noindent where $\{ {{\partial ^T}\over{\partial \vec x}} {\bf \Sigma
 _{\vec x}} {{\partial }\over{\partial \vec x}} \}$ and $\{ {{\partial
 ^T}\over{\partial \vec v}} {\bf \Sigma _{\vec v}} {{\partial
 }\over{\partial \vec v}} \}$ are scalar differential operators, and
 $\left| V \right|$ is the volume of velocity space $\int d \vec v$.

\newpage

\section* {Acknowledgments}

This work was supported by the AFOSR grant F49620-95-1-0265 to
A.L.Y. and N.M.G. Further support to A.L.Y. came from ARPA and the
Office of Naval Research with grant number N00014-95-1-1022. In
addition, N.M.G. received support from grants by the National Eye
Institute \linebreak (EY08921 and EY11170) and the William
A. Kettlewell chair. Finally, we thank the National Eye Institute for
a core grant to Smith-Kettlewell (EY06883).

\newpage

\section*{References}

\begin{description}

 \item Adelson, E.H., and Bergen, J.R. 1985. Spatio-temporal energy
models for the perception of motion.  {\it J. Opt.  Soc. A.}, {\bf
2}, 284-299.

 \item Anstis, S.M., and Ramachandran, V.S. 1987. Visual inertia in
apparent motion. {\it  Vision Res.} {\bf 27}, 755-764.

 \item Ascher, D.  1998. "Human Visual Speed Perception: Psychophysics and
Modeling.''  Program of Cognitive Science. Department of Cognitive and
Linguistic Sciences. Brown University.

 \item Bar-Shalom, Y., and Fortmann, T.E. 1988. {\it Tracking and Data
Association}. Academic Press.

 \item Burr, D.C., Ross, J., and Morrone,
M.C. 1986. Seeing objects in motion. {\it Proc. Royal Soc. Lond. B},
{\bf 227}, 249-265.

 \item Chin, T.M., Karl, W.C., and Willsky, A.S. 1994. Probabilistic
and sequential computation of optical flow using temporal coherence.
{\em IEEE Trans. Image Proc.} {\bf 3}, 773-788.

 \item Clark, J.J., and Yuille, A.L. 1990. {\it Data Fusion for
Sensory Information Processing Systems.} Kluwer Academic
Press, Boston.

 \item Courant, R., and Hilbert, D. 1953. {\it Methods of Mathematical
Physics,} Volume 1. Interscience, New York.

 \item  Cover, T.M.  and J.A. Thomas, J.A. {\bf Elements of Information
Theory}. Wiley Interscience Press. New York. 1991.

 \item Gottsdanker, R.M. 1956. The ability of human operators to
detect acceleration of target motion.  {\it Psych. Bull.}  {\bf 53},
477-487.

\item Grzywacz, N.M., Amthor, F.R. and Mistler, L.A. 1990. 
Applicability of quadratic and threshold models to motion discrimination in the 
rabbit retina.  {\it Biol. Cybern.} {\bf 64}, 41-49.

\item Grzywacz, N.M. and Koch, C. 1987. Functional properties of 
models for direction selectivity in the retina. {\it Synapse}, {\bf 1}, 417-434.

 \item Grzywacz, N.M., Smith,  J.A., and Yuille, A.L. 1989. A
computational framework for visual motion's spatial and temporal
coherence. {\it Proc. IEEE Workshop on Visual Motion},
Irvine.

 \item Grzywacz, N.M., Watamaniuk, S.N.J., and McKee, S.P. 1995.
Temporal coherence theory for the detection and measurement of visual
motion. {\it Vision Res.} {\bf 35}, 3183-3203.

 \item Grzywacz, N.M.  and Yuille, A.L. 1990. ``A model for
the estimate of local image velocity by cells in the visual cortex''.
{\it Proc. Royal Soc. Lond. B},   {\bf 239}, 129-161. 

 \item Hassenstein, B., and Reichardt, W.E. 1956.  Systemtheoretische
analyse der zeit-, reihenfolgen- und vorzeichenauswertung bei der
bewegungsperzeption des russelkafers {\it Chlorophanus}. {\em
Z. Naturforsch.} {\bf 11b}, 513-524.

 \item Heeger, D.J. 1992. Normalization of cell responses in
cat striate cortex. {\it Visual Neurosci.} {\bf 9}, 181-197.

 \item Ho, Y-C., and Lee, R.C.K. 1964. A Bayesian approach to problems
in stochastic estimation and control. {\it IEEE Trans. on Automatic
Control} {\bf 9}, 333-339.

\item Holub, R.A. and Morton-Gibson, M.  1981. Response of
visual cortical neurons of the cat to moving sinusoidal
gratings: Response-contrast functions and spatiotemporal
integration {\it J. Neurophysiol.} {\bf 46}, 1244--1259.

 \item Huber, P.J. 1981. {\it Robust Statistics.} John Wiley and
Sons, New York.

 \item Isard, M., and Blake A. 1996. Contour tracking by stochastic
propagation of conditional density. {\it Proc. Europ. Conf. Comput.
Vision}, pp. 343-356, Cambridge, UK.

 \item Isard, M., and Blake, A. 1998. A mixed-state condensation tracker with automatic
model-switching. {\it Proc. Int. Conf. Comp. Vis.}, pp. 107Ð112, Mumbai, India.
 
 \item Jazwinski, A.H. 1970. {\it Stochastic Processes and Filtering
Theory}, Academic Press, San Diego.

 \item Kalman, R.E. 1960. A new approach to linear filtering and
prediction problems. {\it Trans. ASME, D: J. Basic
Engineering}, {\bf 82}, 35-45.

 \item Koch, C., and Poggio, T. 1992. Multiplying with synapses and
neurons. In {\it Single Neuron Computation}, T. McKenna, J. Davis, and
S.F. Zornetzer eds, pp. 315--345. Academic Press, Boston.

 \item Kulikowski, J. J. and Tolhurst, D. J. 1973.  Psychophysical
evidence for sustained and transient detectors in human vision.  {\it
Journal of Physiology}, {\bf 232}, 519-548.

 \item Matthies, L., Kanade, T., and Szeliski, R. 1989.  Kalman
filter-based algorithms for estimating depth from image sequences.
{\em Int. J. Comput. Vision} {\bf 3}, 209-236.

 \item Maunsell, J. H. R. and Newsome, W. T. 1987. Visual Processing
in Monkey Extrastriate Cortex. In Cowan, W. M., Shooter, E. M.,
Stevens, C. F. and Thompson, R. F. (Eds.) {\it Annual Reviews of
Neuroscience} (vol. 10, pp. 363--401).  Palo Alto, CA: Annual Reviews
Inc.

 \item Mel B.W., Ruderman D.L., Archie K.A. 1988. Translation-invariant orientation tuning in visual
"complex" cells could derive from intradendritic computations. {\it J Neurosci.}, {\bf 18}, 4325-4334 

 \item Mel B.W. 1993. Synaptic integration in an excitable dendritic tree. {\it J. Neurophysiol} {\bf 70}, 1086-1101.  

 \item McKee, S.P., and Welch, L. 1985. Sequential recruitment in the
discrimination of velocity. {\it J. Optical Soc. America A} {\bf 2},
243-251.

 \item McKee, S.P., Silverman, G.H., and Nakayama,
K. 1986. Precise velocity discrimination despite random variations in
temporal frequency and contrast. {\it Vision Res.} {\bf 26}, 609-619.

 \item Merigan WH, Maunsell JH. 1993.  How parallel are the primate
visual pathways?  In Cowan, W. M., Shooter, E. M., Stevens, C. F.  \&
Thompson, R. F. (Eds.) {\it Annual Reviews of Neuroscience} (vol. 16,
pp. 369--4021).  Palo Alto, CA: Annual Reviews Inc.

 \item Nakayama, K. 1985. Biological image motion processing: A
 review. {\it Vision Res.} {\bf 25}, 625-660.

\item Poggio T and Reichardt WE. 1973. Considerations on models 
of movement detection.  {\em Kybernetik}, {\bf 13}, 223-227.

\item Poggio, T. and Reichardt, W.E. 1976. Visual control of 
orientation behaviour in the fly: Part II: Towards the underlying neural 
interactions.  {\em Q. Rev. Biophys.}, {\bf 9}, 377-438.

 \item Ramachandran, V.S., and Anstis, S.M. 1983. Extrapolation of
motion path in human visual perception.  {\em Vision Res.} {\bf 23},
83-85.

\item Reichardt, W. 1961. Autocorrelation, a principle for the
evaluation of sensory information by the nervous system.
In {\em Sensory Communication} Ed Rosenblith, Wiley, New York.

 \item Snowden, R.J., and Braddick, O.J. 1989. Extension of
displacement limits in multiple-exposure sequences of apparent motion.
{\em Vision Res.} {\bf 29}, 1777-1787.

 \item van Santen, J.P.H., and Sperling, G. 1984. A temporal
covariance model of motion perception {\em J. Opt.
Soc. A., A} {\bf 1}, 451-473.

\item Torre, V and Poggio, T. 1978. A synaptic mechanism possibly 
underlying directional selectivity to motion. {\em Proc. R. Soc. Lond. 
B.},  {\bf 202}, 409-416.

 \item Verghese, P., Watamaniuk, S.N.J., McKee, S.P. and Grzywacz,
N.M. 1999.  Local motion detectors cannot account for the
detectability of an extended motion trajectory in noise. {\it Vision Res.}
{\bf 39}, 19-30.

\item Watamaniuk, S. N. J., Sekuler, R. \& Williams, D. W. 1989.
Direction perception in complex dynamic displays: the integration of
direction information. {\it Vision Res.}, {\bf 29}, 47-59.

 \item Watamaniuk, S.N.J., McKee, S.P., and Grzywacz,
N.M. 1994. Detecting a Trajectory Embedded in Random-Direction Motion
Noise. {\it Vision Res.}  {\bf 35}, 65-77.

 \item Watamaniuk, S.N.J., and McKee, S.P. 1995. Seeing motion behind
occluders. {\it Nature} {\bf 377}, 729-730.

 \item Watson, A.B., and Ahumada, A.J. 1985. Model of human visual
motion sensing. {\em J. Opt. Soc. A, A} {\bf 2}, 322-342.

 \item Welch L., MacLeod D.I., Mckee S.P. 1997 Motion interference: perturbing perceived direction.
{\it Vision Res.},  {\bf 37}, 2725-2736.

 \item Werkhoven, P., Snippe, H.P., and Toet, A. 1992. Visual
processing of optic acceleration. {\em Vision Res.} {\bf 32}, 2313-2329.

 \item Williams, L.R., and Jacobs, D.W. 1997a. Local parallel
computation of stochastic completion fields. {\em Neural Comput.}
{\bf 9}, 859-881.

 \item Williams, L.R., and Jacobs, D.W. 1997b. Stochastic completion
fields: a neural model of illusory contour shape and salience. {\em
Neural Comput.} {\bf 9}, 837-858.

 \item Yuille, A.L., Burgi, P.-Y., and Grzywacz, N.M. 1998. Visual
 motion estimation and prediction: A probabilistic network model for
 temporal coherence.  Smith-Kettlewell Eye Research Institute
Technical Report.

 \item Yuille, A.L., Burgi, P.-Y., and Grzywacz, N.M. 1998. Visual
 motion estimation and prediction: A probabilistic network model for
 temporal coherence.  In {\em Proceedings of the Sixth International
 Conference on Computer Vision,} (pp. 973-978). IEEE Computer Society.

\item Yuille, A.L. and Grzywacz, N.M. 1988. ``A Computational Theory
 for the Perception of Coherent Visual Motion". {\it Nature}. {\bf
 333}, pp 71-74.

 \item Yuille, A.L. and Grzywacz, N.M. 1989. A Mathematical Analysis of the Motion
 Coherence Theory.  {\em International Journal of Computer Vision,}
 {\bf 3,} 155-175.  

 \item Yuille, A.L., and Grzywacz, N.M. 1998.  A Theoretical framework
for visual motion. In {\em High-Level Motion Processing :
Computational, Neurophysiological, and Psychophysical
Perspectives}, T. Watanabe ed.  MIT Press.

 \item Yuille, A.L., and Grzywacz, N.M. A computational theory for the
perception of coherent visual motion. {\it Nature}. 333. pp 71-74. 1988.

\end{description}

\end{document}